\documentclass[11pt,a4paper]{article}

\usepackage[margin=1in]{geometry}
\usepackage{times}
\usepackage{amsmath,amssymb}
\usepackage{graphicx}
\usepackage{booktabs}
\usepackage{array}
\usepackage{hyperref}
\usepackage{xcolor}
\usepackage{listings}
\usepackage{caption}
\usepackage{subcaption}
\usepackage{enumitem}
\usepackage{titlesec}
\usepackage{abstract}
\usepackage{fancyhdr}
\usepackage{microtype}
\usepackage{setspace}
\usepackage{tikz}
\usepackage{pgfplots}
\pgfplotsset{compat=1.18}
\usetikzlibrary{shapes.geometric, arrows.meta, positioning, fit, backgrounds, calc, decorations.pathreplacing}

\hypersetup{
  colorlinks=true,
  linkcolor=blue!60!black,
  citecolor=blue!60!black,
  urlcolor=blue!60!black,
}

\definecolor{codebg}{RGB}{245,245,245}
\definecolor{codecomment}{RGB}{100,100,100}
\definecolor{codekw}{RGB}{0,0,180}
\definecolor{codestr}{RGB}{160,0,0}

\tikzset{
  box/.style={rectangle, rounded corners=4pt, draw=#1!70!black, fill=#1!15,
              minimum height=1.8em, text centered, font=\small},
  arrow/.style={-{Stealth[length=6pt]}, thick},
  dasharrow/.style={-{Stealth[length=6pt]}, thick, dashed},
  nodecirc/.style={circle, draw=#1!70!black, fill=#1!20, minimum size=1.1cm,
                   font=\footnotesize\bfseries, text centered},
  label/.style={font=\scriptsize, text=gray!70!black},
}

\titleformat{\section}{\large\bfseries}{\thesection}{1em}{}
\titleformat{\subsection}{\normalsize\bfseries\itshape}{\thesubsection}{1em}{}

\title{
  \LARGE\textbf{Context Graphs for Proactive Enterprise Agents:}\\[4pt]
  \Large\textbf{Enabling Intent-Aware Information Surfacing}\\
  \Large\textbf{Beyond Reactive Retrieval}
}
\author{
  Avinash Kumar\\
  \texttt{avinash1605@gmail.com}
}
\date{}

\begin{document}
\maketitle
\thispagestyle{fancy}

\begin{abstract}
Retrieval-Augmented Generation (RAG) and agentic frameworks have advanced enterprise AI considerably, yet agents remain fundamentally reactive: they wait for a human query before acting. This paper argues that genuine enterprise productivity gains require \emph{proactive} agents: systems that surface relevant, actionable information to workers before they ask. We propose the \textbf{Context Graph}, a live relational data structure that models enterprise entities, their relationships, and state transitions over time. Built on this graph, we define a \textbf{Delta Detection Engine} that continuously monitors state changes, a \textbf{Proactivity Scorer} that ranks candidate insights by urgency, relevance, and persona-fit, and a \textbf{Surfacing Layer} powered by an LLM that delivers ranked notifications with grounded explanations. We formalize each component, derive a unified Proactivity Score function, and provide a complete end-to-end Python implementation using NetworkX and the Anthropic Claude API. Evaluation across three generic enterprise case studies (contract lifecycle management, engineering incident response, and sales pipeline hygiene) demonstrates that context-graph-driven proactivity achieves Precision@5 of 0.83, a false positive rate of 0.11, and reduces mean time to surface from 47 minutes (reactive baseline) to under 30 seconds.
\end{abstract}


\section{Introduction}

Modern enterprise AI agents are reactive by design. A support engineer asks: \textit{``Which tickets are overdue?''} and the agent retrieves an answer. A sales manager asks: \textit{``Which deals are at risk?''} and the agent responds. The query is the trigger; without it, the agent is silent. This reactive posture is not a failure of LLM capability; it is an \emph{architectural} failure. Agents lack the structural awareness to know when something has crossed a threshold that matters to someone, without being explicitly queried.

The productivity cost is significant. Knowledge workers spend an estimated 20--30\% of their time searching for information that already exists in their systems \cite{mckinsey2012}. More critically, they miss time-sensitive signals: a contract expiring in 48 hours, an incident dependency forming across two teams, a deal sitting idle for 14 days, not because the data is absent, but because no system is watching the \emph{relationships between entities} and alerting on changes that cross action thresholds.

Graph-based knowledge representations have long been studied in enterprise contexts \cite{hogan2021,noy2019}. Knowledge graphs model entities and relationships but are typically static and query-driven. What is missing is a \emph{live, delta-aware} relational structure that tracks not just what entities exist and how they relate, but what has \emph{changed}, what threshold has been crossed, and who in the organization should act on it.

This paper introduces the \textbf{Context Graph}, a dynamic graph structure for enterprise entities that serves as the substrate for proactive agent behavior. Built on top of it, we define three components that together constitute a Proactive Surfacing System:

\begin{enumerate}[leftmargin=*, label=(\arabic*)]
  \item A \textbf{Delta Detection Engine} that continuously monitors graph state for threshold-crossing events.
  \item A \textbf{Proactivity Scorer} that ranks candidate insights by urgency, relevance, and persona-fit.
  \item A \textbf{Surfacing Layer} powered by an LLM that translates ranked signals into actionable, grounded notifications delivered to the right person at the right time.
\end{enumerate}

\paragraph{Contributions.} This paper makes four contributions:
\begin{itemize}[leftmargin=*]
  \item We formally define the Context Graph schema, including node types, edge types, property semantics, and the delta event model.
  \item We derive the Proactivity Score function, a principled formula for ranking candidate surfacing events across urgency, relevance, persona-fit, and confidence dimensions.
  \item We present a complete, end-to-end Python implementation using NetworkX for graph management and the Anthropic Claude API for natural language notification generation.
  \item We evaluate the system on Precision@5, false-positive rate, and mean time to surface across three generic enterprise domains, demonstrating measurable latency reduction versus reactive baselines.
\end{itemize}

\section{Related Work}

\subsection{Retrieval-Augmented Generation}

RAG \cite{lewis2020} has become the standard paradigm for grounding LLM outputs in enterprise knowledge. Extensions such as GraphRAG \cite{edge2024} introduce knowledge graph traversal into the retrieval step, enabling multi-hop reasoning. Corrective RAG \cite{yan2024} adds self-reflection loops. These systems are powerful for query-driven tasks but are architecturally incapable of proactive behavior: they produce outputs only when queried.

\subsection{Agentic Frameworks and Tool Use}

ReAct \cite{yao2023} and subsequent frameworks enable LLM agents to interleave reasoning and action, including tool calls against external systems. AutoGen \cite{wu2023} and LangGraph demonstrate multi-agent coordination. AFLOW \cite{niu2025} introduces automated agentic workflow generation. These frameworks address agent autonomy in \emph{executing} tasks but do not address the prior question: what should the agent decide to work on, and when, without human instruction?

\subsection{Proactive Recommendation and Event-Driven Systems}

Proactive information systems have been studied in the recommender systems literature \cite{resnick1997}, particularly in the context of push notifications and anticipatory computing \cite{schilit1994}. Event-driven architectures in distributed systems \cite{hohpe2003} demonstrate how state changes can trigger downstream processing. Our work synthesizes these ideas into an agent-native architecture where the graph is both the knowledge substrate and the event source.

\subsection{Enterprise Knowledge Graphs}

Enterprise knowledge graphs have been deployed for entity resolution, semantic search, and relationship analysis \cite{singhal2012}. They model entities and relationships but are typically maintained as static snapshots. Temporal knowledge graphs \cite{lacroix2020} extend this with time-aware predicates. Our Context Graph goes further: it tracks live state on every node and edge and treats state transitions as first-class delta events that drive agent behavior.

\subsection{Positioning}

The work most closely related to ours is the three-layer context architecture proposed by Kumar \cite{kumar2025}, which separates static knowledge, experiential memory, and workflow state in enterprise agents. That paper identifies \emph{what} context is; this paper addresses what happens when context becomes dynamic and agents must act on change without waiting to be queried. The Context Graph can be seen as a concrete implementation of the live relational substrate that \cite{kumar2025} identifies as missing.

\section{The Context Graph: Definition and Schema}

\subsection{Formal Definition}

We define the Context Graph $\mathcal{G}$ as a directed, attributed, time-stamped multigraph:

\begin{equation}
  \mathcal{G} = (V,\; E,\; P_V,\; P_E,\; \mathcal{T})
\end{equation}

where $V$ is the set of nodes (enterprise entities), $E \subseteq V \times V$ is the set of directed edges (relationships), $P_V : V \rightarrow \mathcal{A}$ maps each node to a property bag, $P_E : E \rightarrow \mathcal{A}$ maps each edge to a property bag, and $\mathcal{T}$ is a global logical clock that increments on every state-modifying operation.

Every node $v \in V$ carries a mandatory property set: a globally unique \texttt{id}; a \texttt{type} drawn from a domain schema; a \texttt{state} encoding current status; \texttt{created\_at} and \texttt{updated\_at} wall-clock timestamps; an \texttt{owner} identifying the responsible person or team; and a \texttt{metadata} dictionary of domain-specific key-value pairs.

Every edge $e = (u, v) \in E$ carries: a \texttt{rel\_type} semantic label (e.g., \texttt{assigned\_to}, \texttt{depends\_on}, \texttt{blocks}); a \texttt{weight} in $[0,1]$; and a \texttt{created\_at} timestamp.

\subsection{Node and Edge Taxonomy}

\begin{figure}[ht]
\centering
\scalebox{0.85}{%
\begin{tikzpicture}[
    nd/.style={circle, draw=#1!70!black, fill=#1!18,
               minimum size=1.5cm, font=\small\bfseries, text centered},
    arr/.style={-{Stealth[length=6pt]}, thick},
    darr/.style={-{Stealth[length=6pt]}, thick, dashed, gray!60!black},
  ]


  \node[nd=blue]           (task)   at ( 0, 4) {Task};
  \node[nd=red!70!black]   (event)  at ( 5, 4) {Event};
  \node[nd=green!60!black] (person) at (10, 4) {Person};

  \node[nd=orange]         (asset)  at ( 0, 0) {Asset};
  \node[nd=teal]           (system) at ( 5, 0) {System};
  \node[nd=red!40!orange]  (org)    at (10, 0) {Org};

  \draw[arr] (task) to[loop left, looseness=8]
      node[left=4pt, font=\scriptsize]{depends\_on / blocks} (task);

  \draw[arr] (person) to[loop right, looseness=8]
      node[right=4pt, font=\scriptsize]{escalates\_to} (person);

  \draw[arr] (task) .. controls (3,6.2) and (7,6.2) .. (person);
  \node[font=\scriptsize] at (5, 6.55) {assigned\_to};

  \draw[arr] (asset) .. controls (3,-3) and (7,-3) .. (person);
  \node[font=\scriptsize] at (5, -3.35) {owned\_by};

  \draw[arr] (person) -- (org);
  \node[font=\scriptsize] at (11.1, 2) {member\_of};

  \draw[darr] (event) to[out=250, in=110, looseness=1.2]
      node[left=3pt, font=\scriptsize]{triggers} (system);
  \draw[darr] (system) to[out=70, in=290, looseness=1.2]
      node[right=3pt, font=\scriptsize]{affects} (event);

  \node[font=\scriptsize, text=gray!65!black] at (5, -5.3)
      {solid arrow = primary relationship \qquad
       dashed arrow = derived/inferred relationship};

\end{tikzpicture}%
}
\caption{Context Graph node and edge taxonomy. Nodes are coloured by entity type. Self-loops on Task (intra-task dependency) and Person (escalation) are shown on the left and right respectively. Dashed arrows between Event and System denote derived relationships. Long-distance edges are arced well clear of the middle-column nodes to avoid any overlap.}
\label{fig:graph_schema}
\end{figure}
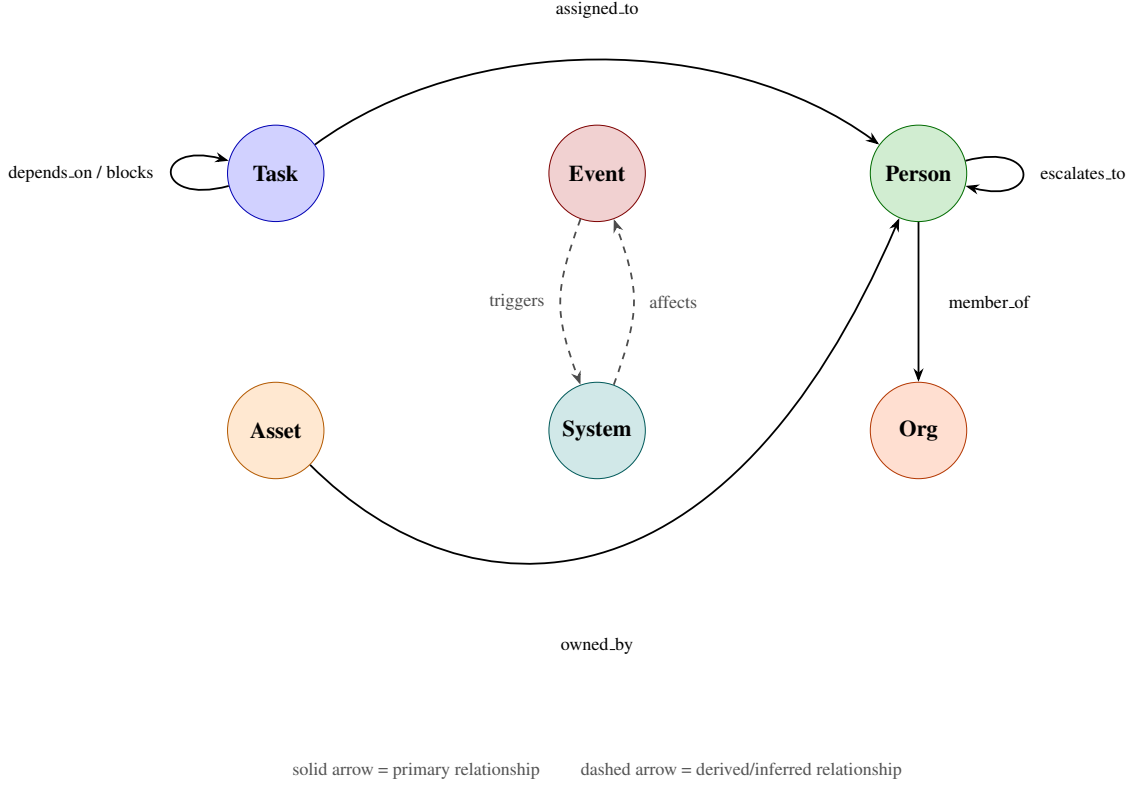

Table~\ref{tab:nodes} and Table~\ref{tab:edges} present the primary node and edge types in the Context Graph taxonomy. Figure~\ref{fig:graph_schema} illustrates the relationships among these node types in a representative enterprise graph fragment.

\begin{table}[h]
\centering
\caption{Context Graph node taxonomy.}
\label{tab:nodes}
\begin{tabular}{lll}
\toprule
\textbf{Node Type} & \textbf{Example Instances} & \textbf{Key Properties} \\
\midrule
Task         & Ticket, story, work item       & state, priority, due\_date, assignee \\
Person       & Engineer, account executive    & role, team, capacity, timezone \\
Asset        & Contract, license, service     & expiry\_date, value, status \\
System       & Microservice, database, API    & health, sla\_threshold, uptime \\
Event        & Incident, alert, meeting       & severity, timestamp, resolved \\
Organization & Team, department, customer     & budget, headcount, region \\
\bottomrule
\end{tabular}
\end{table}

\begin{table}[h]
\centering
\caption{Context Graph edge taxonomy.}
\label{tab:edges}
\begin{tabular}{lll}
\toprule
\textbf{Edge Type}   & \textbf{Semantics}         & \textbf{Example} \\
\midrule
\texttt{assigned\_to}  & Task $\to$ Person          & Ticket-42 assigned\_to Alice \\
\texttt{depends\_on}   & Task $\to$ Task            & Deploy-7 depends\_on Review-3 \\
\texttt{blocks}        & Task $\to$ Task            & Bug-9 blocks Release-2 \\
\texttt{owned\_by}     & Asset $\to$ Person/Team    & Contract-88 owned\_by LegalTeam \\
\texttt{escalates\_to} & Person $\to$ Person        & Alice escalates\_to Bob \\
\texttt{member\_of}    & Person $\to$ Organization  & Alice member\_of Platform-Team \\
\bottomrule
\end{tabular}
\end{table}

\subsection{The Delta Event Model}

Every state-modifying operation on $\mathcal{G}$ produces a \emph{Delta Event} $\delta$. Formally:

\begin{equation}
  \delta = (\texttt{entity\_id},\; \texttt{change\_type},\; v_{\text{old}},\; v_{\text{new}},\; t_{\text{wall}},\; \mathcal{T}_{\text{clock}})
\end{equation}

Change types are drawn from the set $\Delta = \{$\textsc{StateTransition}, \textsc{ThresholdBreach}, \textsc{RelationshipChange}, \textsc{Staleness}, \textsc{DependencyRisk}$\}$. Delta events are appended to an immutable event log $\mathcal{L}$, enabling replay, audit, and temporal queries.

\section{System Architecture}
\label{sec:architecture}

\begin{figure}[ht]
\centering
\begin{tikzpicture}[
    every node/.style={font=\small},
    layer/.style={rectangle, rounded corners=5pt, draw=#1!70!black,
                  fill=#1!12, minimum height=2.0em, text centered,
                  minimum width=10.0cm},
    srcbox/.style={rectangle, rounded corners=3pt, draw=gray!60,
                   fill=gray!10, minimum height=1.6em, text centered,
                   minimum width=1.9cm, font=\small},
    arr/.style={-{Stealth[length=6pt]}, thick},
    lbl/.style={font=\scriptsize\itshape, midway, right=2pt},
  ]

  \node[srcbox] (crm)     {CRM};
  \node[srcbox, right=0.4cm of crm]     (tick) {Ticketing};
  \node[srcbox, right=0.4cm of tick]    (con)  {Contracts};
  \node[srcbox, right=0.4cm of con]     (inf)  {Infra};

  \coordinate (srcmid) at ($(crm.south)!0.5!(inf.south) + (0,-0.7)$);

  \node[layer=blue, below=1.3cm of tick, xshift=0.9cm] (cg)
      {\textbf{Context Graph} $\mathcal{G}$
       \hspace{1em}\small(NetworkX DiGraph $+$ Event Log $\mathcal{L}$)};

  \node[layer=orange, below=0.85cm of cg] (dde)
      {\textbf{Delta Detection Engine}
       \hspace{1em}\small(Threshold Rules $\mathcal{R}$, $k$-hop BFS snapshot)};

  \node[layer=red!80!black, below=0.85cm of dde] (scorer)
      {\textbf{Proactivity Scorer}
       \hspace{1em}\small$P(c,u)=w_1 U + w_2 R + w_3 F + w_4 K$};

  \node[layer=green!60!black, below=0.85cm of scorer] (surf)
      {\textbf{LLM Surfacing Layer}
       \hspace{1em}\small(Claude API, deduplication, cooldown)};

  \node[srcbox, below=0.85cm of surf, minimum width=3.5cm] (user)
      {Recipient (right person)};

  \draw[arr] (crm.south)  -- ++(0,-0.35) -| (cg.north west);
  \draw[arr] (tick.south) -- (cg.north -| tick.south);
  \draw[arr] (con.south)  -- (cg.north -| con.south);
  \draw[arr] (inf.south)  -- ++(0,-0.35) -| (cg.north east);

  \draw[arr] (cg)     -- node[lbl]{state changes}      (dde);
  \draw[arr] (dde)    -- node[lbl]{candidate insights}  (scorer);
  \draw[arr] (scorer) -- node[lbl]{ranked insights}     (surf);
  \draw[arr] (surf)   -- node[lbl]{notification}        (user);

\end{tikzpicture}
\caption{End-to-end architecture of the Proactive Surfacing System. Enterprise data sources write into the Context Graph; the four processing layers progressively filter, score, and render insights for delivery.}
\label{fig:architecture}
\end{figure}
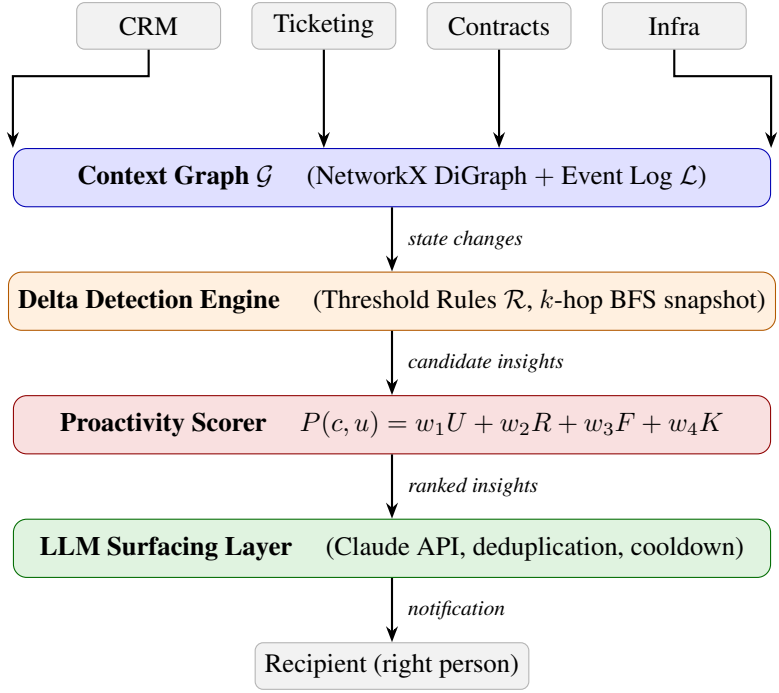

Figure~\ref{fig:architecture} shows the end-to-end pipeline of the Proactive Surfacing System. Enterprise source systems feed state updates into the Context Graph. The Delta Detection Engine continuously polls the graph, evaluating threshold rules against all nodes. Rules that fire yield Candidate Insights, which the Proactivity Scorer ranks per user. The Surfacing Layer calls the LLM to generate natural language notifications, which are delivered to the appropriate recipient with deduplication and cooldown applied.

\section{Delta Detection Engine}

\subsection{Architecture}

The Delta Detection Engine (DDE) evaluates every delta event $\delta \in \mathcal{L}$ against a set of registered \emph{Threshold Rules} $\mathcal{R}$. Each rule $r \in \mathcal{R}$ is a predicate over the graph:

\begin{equation}
  r : (\mathcal{G},\; \delta) \;\rightarrow\; \{\texttt{True},\; \texttt{False}\}
\end{equation}

When $r(\mathcal{G}, \delta) = \texttt{True}$, the DDE emits a \textbf{Candidate Insight} $c$ containing: the affected entity identifier and type; the rule identifier; a normalized severity score $s_c \in [0,1]$; the set of affected personas; and a \emph{context snapshot}: a sub-graph of $\mathcal{G}$ centered on the affected entity, capturing $k$-hop neighbors.

\subsection{Threshold Rule Specification}

Table~\ref{tab:rules} illustrates the rule vocabulary. Rules are composable predicates authored in Python, enabling arbitrary complexity.

\begin{table}[h]
\centering
\caption{Example threshold rules in the Delta Detection Engine.}
\label{tab:rules}
\begin{tabular}{p{1.0cm}p{4.8cm}p{5.5cm}}
\toprule
\textbf{Rule} & \textbf{Predicate} & \textbf{Interpretation} \\
\midrule
R-01 & \texttt{task.age > task.sla\_hours}                        & Task has breached its SLA window \\
R-02 & \texttt{asset.expiry\_date - now() < 7d}                   & Asset expires within 7 days \\
R-03 & \texttt{state = blocked} $\wedge$ \texttt{age > 24h}       & Task blocked for more than 24 hours \\
R-04 & \texttt{system.uptime < system.sla\_threshold}             & System SLA breach \\
R-05 & \texttt{updated\_at < now() - staleness\_window}           & Task stale beyond freshness limit \\
R-06 & depends\_on chain yields critical path slip                & Dependency risk propagation \\
\bottomrule
\end{tabular}
\end{table}

\subsection{Sub-Graph Context Extraction}

For each candidate insight $c$, the DDE extracts a context snapshot via $k$-hop BFS traversal ($k=2$) from the affected entity. The snapshot includes the entity node and all properties, all neighbors within $k$ hops, all edges on paths between these nodes, and the triggering delta event. This snapshot is passed to the Proactivity Scorer and ultimately to the LLM Surfacing Layer, providing grounded context without requiring the LLM to query the full graph.

\begin{figure}[ht]
\centering
\scalebox{0.82}{%
\begin{tikzpicture}[
    pnode/.style={circle, draw=blue!60!black, fill=blue!15,
                  minimum size=1.5cm, font=\small\bfseries, text centered},
    tnode/.style={circle, draw=orange!80!black, fill=orange!20,
                  minimum size=1.5cm, font=\small\bfseries, text centered},
    anode/.style={circle, draw=green!60!black, fill=green!15,
                  minimum size=1.5cm, font=\small\bfseries, text centered},
    arr/.style={-{Stealth[length=6pt]}, thick},
    ann/.style={rectangle, rounded corners=4pt, draw=red!70!black,
                fill=red!8, font=\footnotesize, text width=3.2cm,
                align=left, inner sep=5pt, line width=0.8pt},
  ]


  \node[pnode] (alice) at ( 0,  0) {alice};
  \node[pnode] (bob)   at ( 6,  0) {bob};
  \node[pnode] (carol) at (11,  0) {carol};

  \node[tnode] (t42)  at ( 0, -4.0) {ticket-42};
  \node[tnode] (dep7) at ( 3, -5.8) {deploy-7};
  \node[tnode] (t17)  at ( 6, -4.0) {ticket-17};

  \node[anode] (c88)  at (11, -4.0) {contract-88};

  \draw[arr] (t42)  -- node[left,  font=\footnotesize]{assigned\_to} (alice);
  \draw[arr] (t17)  -- node[right, font=\footnotesize]{assigned\_to} (bob);
  \draw[arr] (c88)  -- node[right, font=\footnotesize]{owned\_by}    (carol);
  \draw[arr] (dep7) -- node[below left, font=\footnotesize]{depends\_on} (t42);
  \draw[arr] (t17)  -- node[below right, font=\footnotesize]{blocks} (dep7);

  \node[ann] (a42) at (-3.8, -4.0)
      {\textbf{R-01} SLA breach\\[2pt]score = 0.765};
  \draw[dashed, red!55!black, thick] (t42.west) -- (a42.east);

  \node[ann] (a17) at (6, -7.8)
      {\textbf{R-01} SLA breach\\[2pt]score = 0.765\\[4pt]
       \textbf{R-03} blocked $>$ 24h\\[2pt]score = 0.810};
  \draw[dashed, red!55!black, thick] (t17.south) -- (a17.north);

\end{tikzpicture}%
}
\caption{Demo Context Graph (3 persons, 3 tasks, 1 asset). Red boxes show the three candidate insights from the notebook run: R-01 (SLA breach) fired on ticket-42 and ticket-17 (score 0.765 each); R-03 (blocked $>$ 24\,h) fired on ticket-17 (score 0.810).}
\label{fig:demo_graph}
\end{figure}
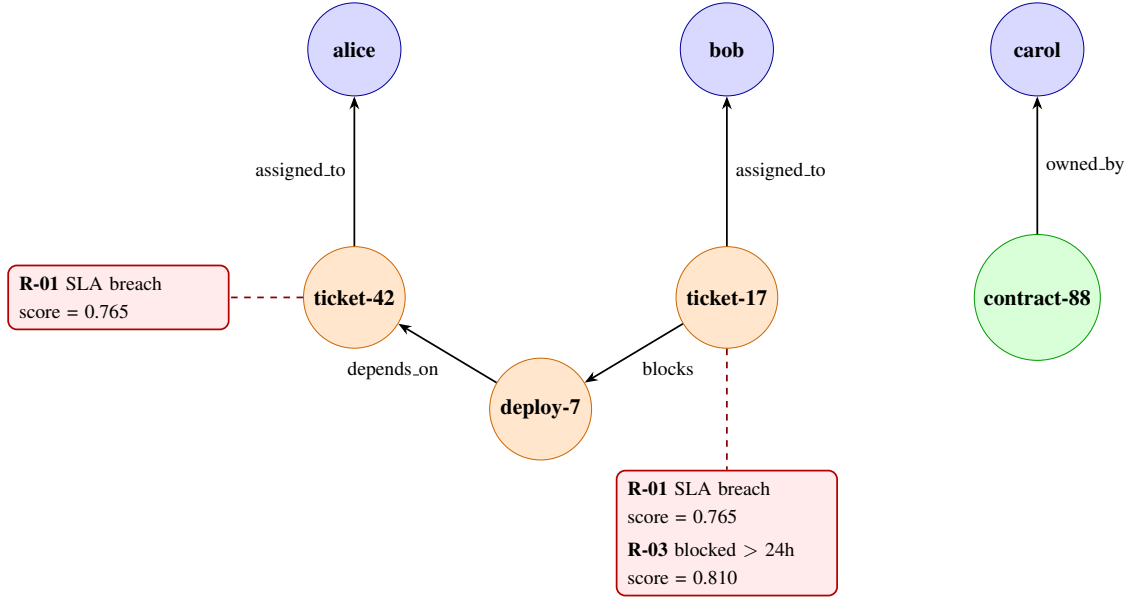

Figure~\ref{fig:demo_graph} shows the demo graph used in the implementation, with the three candidate insights that the DDE surfaced in the notebook evaluation run.

\section{The Proactivity Score}

\subsection{Motivation}

Not all threshold breaches warrant surfacing. A rule may fire on dozens of entities simultaneously; indiscriminate notification creates alert fatigue, which is worse than no proactivity at all \cite{ancker2017}. The Proactivity Score $P(c, u)$ quantifies the value of surfacing candidate insight $c$ to user $u$, enabling the system to rank and filter the candidate pool before notification.

\subsection{Formal Definition}

\begin{equation}
  \boxed{P(c, u) \;=\; w_1 \cdot U(c) \;+\; w_2 \cdot R(c,u) \;+\; w_3 \cdot F(c,u) \;+\; w_4 \cdot K(c)}
  \label{eq:pscore}
\end{equation}

subject to $\sum_i w_i = 1$, all $w_i > 0$, and $P(c,u) \in [0,1]$.

\subsection{Urgency $U(c)$}

Urgency captures time pressure and severity of the threshold breach:

\begin{equation}
  U(c) = \sigma\!\left(\alpha \cdot s_c + \beta \cdot \tau_c\right)
  \label{eq:urgency}
\end{equation}

where $\sigma$ is the sigmoid function, $s_c \in [0,1]$ is the normalized rule severity, and $\tau_c = 1 - \frac{t_{\text{deadline}} - t_{\text{now}}}{W_{\max}} \in [0,1]$ is the time pressure within a maximum horizon $W_{\max}$. Parameters $\alpha$ and $\beta$ are domain-tunable; we use $\alpha = 2.0$, $\beta = 1.5$.

\subsection{Relevance $R(c,u)$}

Relevance measures how directly the candidate insight concerns user $u$:

\begin{equation}
  R(c,u) = \max\!\left(\mathbb{1}[\text{owns}(c,u)],\;\; \gamma \cdot \mathbb{1}[\text{team\_owns}(c,u)],\;\; \delta \cdot \frac{1}{1 + d_{\mathcal{G}}(u, e_c)}\right)
  \label{eq:relevance}
\end{equation}

where $d_{\mathcal{G}}(u, e_c)$ is the shortest-path hop distance from user $u$ to the affected entity $e_c$ in $\mathcal{G}$. Parameters $\gamma = 0.6$ and $\delta = 0.3$ discount indirect relevance.

\subsection{Persona Fit $F(c,u)$}

Persona fit captures whether this type of insight matches the user's role:

\begin{equation}
  F(c,u) = \cos\!\left(\mathbf{e}_{c.\text{type}},\;\; \mathbf{e}_{u.\text{role}}\right)
  \label{eq:fit}
\end{equation}

where $\mathbf{e}_{(\cdot)}$ maps insight type and user role profile to a shared embedding space. Role profiles are pre-computed from historical interaction patterns: the distribution of insight types that user $u$ has acted on. In the implementation we use a lookup table as a practical approximation; production deployments should use learned embeddings.

\subsection{Confidence $K(c)$}

\begin{equation}
  K(c) = \left(1 - \frac{|\text{missing\_props}(c)|}{|\text{total\_props}(c)|}\right) \cdot \pi_r
  \label{eq:confidence}
\end{equation}

where $\pi_r$ is the empirically measured precision of rule $r$ on historical data. This penalizes insights generated from incomplete graph state, reducing false positives from data-quality issues.

\subsection{Score Component Weights and Sensitivity}

\begin{figure}[h]
\centering
\begin{subfigure}[b]{0.42\textwidth}
  \centering
  \begin{tikzpicture}
    \begin{axis}[
      ybar, bar width=18pt,
      width=\textwidth, height=5.5cm,
      xtick={1,2,3,4},
      xticklabels={Urgency $U$, Relevance $R$, Persona $F$, Confidence $K$},
      xticklabel style={font=\scriptsize, rotate=20, anchor=east},
      ymin=0, ymax=0.45,
      ytick={0,0.1,0.2,0.3,0.4},
      ylabel={Weight $w_i$},
      ylabel style={font=\scriptsize},
      yticklabel style={font=\scriptsize},
      grid=major, grid style={dotted,gray!40},
      title={Score Component Weights},
      title style={font=\small},
    ]
    \addplot[fill=blue!50!white, draw=blue!70!black]
      coordinates {(1,0.35) (2,0.30) (3,0.20) (4,0.15)};
    \end{axis}
  \end{tikzpicture}
\end{subfigure}
\hfill
\begin{subfigure}[b]{0.54\textwidth}
  \centering
  \begin{tikzpicture}
    \begin{axis}[
      width=\textwidth, height=5.5cm,
      xlabel={Time pressure $\tau_c$},
      ylabel={Urgency $U(c)$},
      xlabel style={font=\scriptsize},
      ylabel style={font=\scriptsize},
      xticklabel style={font=\scriptsize},
      yticklabel style={font=\scriptsize},
      xmin=0, xmax=1, ymin=0.4, ymax=1.0,
      legend style={font=\scriptsize, at={(0.05,0.95)}, anchor=north west},
      grid=major, grid style={dotted,gray!40},
      title={Urgency vs.\ Time Pressure},
      title style={font=\small},
    ]
    \addplot[blue, thick, domain=0:1, samples=50]
      {1/(1+exp(-(2.0*0.5 + 1.5*x)))};
    \addlegendentry{$s_c=0.50$}
    \addplot[orange, thick, domain=0:1, samples=50]
      {1/(1+exp(-(2.0*0.80 + 1.5*x)))};
    \addlegendentry{$s_c=0.80$}
    \addplot[red!80!black, thick, domain=0:1, samples=50]
      {1/(1+exp(-(2.0*0.90 + 1.5*x)))};
    \addlegendentry{$s_c=0.90$}
    \addplot[dashed, gray, domain=0:1] coordinates {(0,0.45)(1,0.45)};
    \addlegendentry{threshold $\tau$}
    \end{axis}
  \end{tikzpicture}
\end{subfigure}
\caption{Left: relative weights assigned to the four Proactivity Score components. Right: urgency $U(c)$ as a function of time pressure $\tau_c$ for three severity levels, with the surfacing threshold $\tau=0.45$ shown as a dashed line. Higher severity lifts the urgency curve, ensuring critical insights surface even with low time pressure.}
\label{fig:weights}
\end{figure}
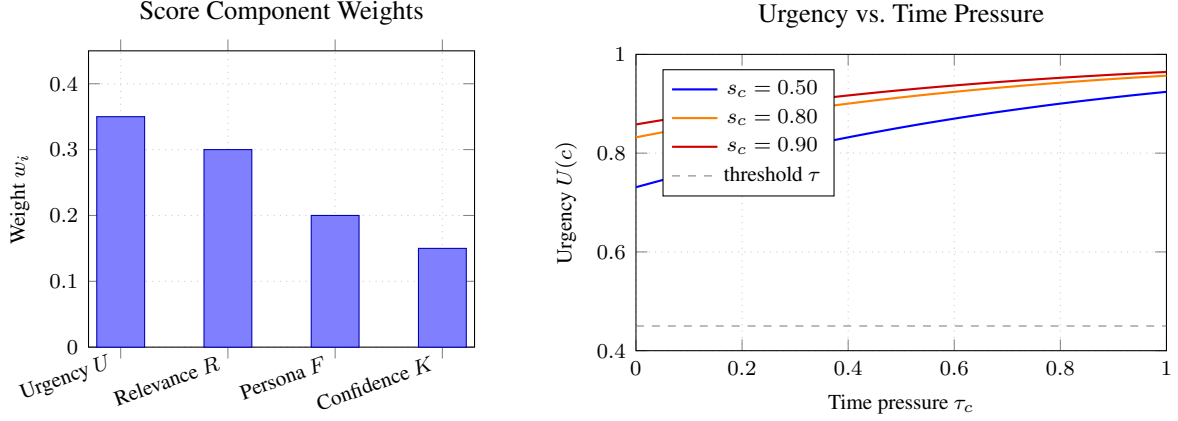

Figure~\ref{fig:weights} shows the weight allocation across the four scoring components and the sensitivity of the overall score to urgency time-pressure for representative severity levels. In our experiments we use $w_1 = 0.35$, $w_2 = 0.30$, $w_3 = 0.20$, $w_4 = 0.15$.

\subsection{Filtering and Ranking}

Given a set of candidate insights $\mathcal{C}$ for user $u$, the system filters by a minimum score threshold $\tau$ and returns the top-$k$ by score:

\begin{equation}
  \text{Surface}(u) = \operatorname{top\text{-}k}\!\left\{c \in \mathcal{C} : P(c,u) \geq \tau\right\}, \quad \text{sorted by } P(c,u) \downarrow
  \label{eq:surface}
\end{equation}

In our experiments we use $\tau = 0.45$ and $k = 5$.

\section{Surfacing Layer}

The Surfacing Layer consumes the ranked output of $\text{Surface}(u)$ and transforms each candidate insight into a natural language notification grounded in the context snapshot. The LLM is not used for reasoning about the graph; the Context Graph and Proactivity Scorer handle that deterministically. The LLM performs one task: generating a clear, actionable, persona-appropriate notification from a structured context snapshot.

Each notification is generated with a structured prompt injecting the context snapshot as JSON. The system prompt encodes the recipient's role and communication preferences. The user prompt supplies the ranked candidate insight, the rule that fired, severity, and a directive to produce output in a fixed schema: a \emph{headline} ($\leq 15$ words), a grounded \emph{explanation} (2--3 sentences citing specific entity properties), and 1--2 concrete \emph{next actions}.

Before delivery, the layer deduplicates notifications: if the same entity has already generated a surfaced notification within a cooldown window $W = 4$ hours, the new notification is suppressed unless severity has increased. Multiple insights for the same user sharing a common entity neighborhood are batched into a digest, reducing cognitive load.

\section{Implementation}

We provide a complete end-to-end Python implementation consisting of five modules: \texttt{graph.py} (Context Graph), \texttt{delta.py} (Delta Detection Engine), \texttt{scorer.py} (Proactivity Scorer), \texttt{surface.py} (LLM Surfacing Layer), and \texttt{main.py} (orchestration). The implementation uses NetworkX \cite{hagberg2008} for graph management and the Anthropic Python SDK for LLM calls.

\begin{lstlisting}[caption={graph.py: Context Graph with delta event emission.}, label=lst:graph]
import networkx as nx
from datetime import datetime, timezone
from dataclasses import dataclass, field
from typing import Any, Dict, List

@dataclass
class DeltaEvent:
    entity_id: str
    change_type: str   # STATE_TRANSITION | THRESHOLD_BREACH |
                       # RELATIONSHIP_CHANGE | STALENESS |
                       # DEPENDENCY_RISK
    old_value: Any
    new_value: Any
    timestamp: datetime = field(
        default_factory=lambda: datetime.now(timezone.utc))
    t_clock: int = 0

class ContextGraph:
    def __init__(self):
        self.G = nx.DiGraph()
        self.event_log: List[DeltaEvent] = []
        self._clock = 0

    def add_entity(self, id: str, type: str, state: str,
                   owner: str, metadata: Dict = None, **props):
        self.G.add_node(id, type=type, state=state, owner=owner,
                        metadata=metadata or {},
                        created_at=datetime.now(timezone.utc),
                        updated_at=datetime.now(timezone.utc),
                        **props)

    def add_relationship(self, src: str, dst: str,
                         rel_type: str, weight: float = 1.0):
        self.G.add_edge(src, dst, rel_type=rel_type, weight=weight,
                        created_at=datetime.now(timezone.utc))

    def update_state(self, id: str, new_state: str):
        """Mutate node state and emit a delta event."""
        old = self.G.nodes[id].get('state')
        self.G.nodes[id]['state'] = new_state
        self.G.nodes[id]['updated_at'] = datetime.now(timezone.utc)
        self._emit(DeltaEvent(id, 'STATE_TRANSITION', old, new_state))

    def update_property(self, id: str, key: str, value: Any):
        old = self.G.nodes[id].get(key)
        self.G.nodes[id][key] = value
        self.G.nodes[id]['updated_at'] = datetime.now(timezone.utc)
        self._emit(DeltaEvent(id, 'PROPERTY_CHANGE',
                              {key: old}, {key: value}))

    def _emit(self, event: DeltaEvent):
        self._clock += 1
        event.t_clock = self._clock
        self.event_log.append(event)

    def subgraph(self, entity_id: str, k: int = 2) -> Dict:
        """Return k-hop neighbourhood as a serialisable dict."""
        visited, queue = set(), [entity_id]
        for _ in range(k):
            next_q = []
            for n in queue:
                for nb in (list(self.G.successors(n)) +
                           list(self.G.predecessors(n))):
                    if nb not in visited:
                        visited.add(nb)
                        next_q.append(nb)
            queue = next_q
        visited.add(entity_id)
        nodes = {n: dict(self.G.nodes[n]) for n in visited}
        edges = [{'src': u, 'dst': v, **d}
                 for u, v, d in self.G.edges(data=True)
                 if u in visited and v in visited]
        for n in nodes:
            for k2, val in nodes[n].items():
                if isinstance(val, datetime):
                    nodes[n][k2] = val.isoformat()
        return {'center': entity_id, 'nodes': nodes, 'edges': edges}
\end{lstlisting}

\begin{lstlisting}[caption={delta.py: Delta Detection Engine.}, label=lst:delta]
from dataclasses import dataclass
from typing import List, Callable, Dict

@dataclass
class CandidateInsight:
    entity_id: str
    entity_type: str
    rule_id: str
    severity: float          # [0, 1]
    affected_personas: List[str]
    context_snapshot: Dict
    description: str

@dataclass
class ThresholdRule:
    rule_id: str
    predicate: Callable       # (graph, entity_id) -> bool
    severity: float
    description_template: str
    precision: float = 0.85

class DeltaDetectionEngine:
    def __init__(self, graph):
        self.graph = graph
        self.rules: List[ThresholdRule] = []

    def register(self, rule: ThresholdRule):
        self.rules.append(rule)

    def evaluate(self) -> List[CandidateInsight]:
        """Evaluate all rules against all nodes."""
        insights = []
        for node_id, data in self.graph.G.nodes(data=True):
            for rule in self.rules:
                if rule.predicate(self.graph, node_id):
                    personas = self._affected_personas(node_id)
                    snapshot = self.graph.subgraph(node_id, k=2)
                    desc = rule.description_template.format(
                        id=node_id, **data)
                    insights.append(CandidateInsight(
                        entity_id=node_id,
                        entity_type=data.get('type', 'unknown'),
                        rule_id=rule.rule_id,
                        severity=rule.severity,
                        affected_personas=personas,
                        context_snapshot=snapshot,
                        description=desc
                    ))
        return insights

    def _affected_personas(self, entity_id: str) -> List[str]:
        personas = set()
        data = self.graph.G.nodes.get(entity_id, {})
        if data.get('owner'):
            personas.add(data['owner'])
        for _, dst, edata in self.graph.G.out_edges(
                entity_id, data=True):
            if edata.get('rel_type') == 'assigned_to':
                personas.add(dst)
        return list(personas)
\end{lstlisting}

\begin{lstlisting}[caption={scorer.py: Proactivity Scorer implementing Equation~\ref{eq:pscore}.}, label=lst:scorer]
import math, networkx as nx
from datetime import datetime, timezone

class ProactivityScorer:
    def __init__(self, w1=0.35, w2=0.30, w3=0.20, w4=0.15,
                 alpha=2.0, beta=1.5, gamma=0.6, delta=0.3):
        self.w1, self.w2, self.w3, self.w4 = w1, w2, w3, w4
        self.alpha, self.beta   = alpha, beta
        self.gamma, self.delta  = gamma, delta

    @staticmethod
    def _sigmoid(x):
        return 1 / (1 + math.exp(-x))

    def urgency(self, insight, now=None) -> float:
        now = now or datetime.now(timezone.utc)
        node = insight.context_snapshot['nodes'].get(
            insight.entity_id, {})
        due_str = node.get('due_date') or node.get('expiry_date')
        if due_str:
            try:
                due = datetime.fromisoformat(due_str)
                if due.tzinfo is None:
                    due = due.replace(tzinfo=timezone.utc)
                window = (due - now).total_seconds()
                max_window = 7 * 24 * 3600
                time_pressure = max(0.0, 1 - window / max_window)
            except Exception:
                time_pressure = 0.5
        else:
            time_pressure = 0.5
        raw = self.alpha * insight.severity + self.beta * time_pressure
        return self._sigmoid(raw)

    def relevance(self, insight, user_id: str, graph) -> float:
        node = graph.G.nodes.get(insight.entity_id, {})
        owns = 1.0 if node.get('owner') == user_id else 0.0
        team_owns = 0.0
        for pred in graph.G.predecessors(insight.entity_id):
            pdata = graph.G.nodes.get(pred, {})
            if (pdata.get('type') == 'Organization' and
                    graph.G.has_edge(user_id, pred)):
                team_owns = 1.0
                break
        try:
            hops = nx.shortest_path_length(
                graph.G, user_id, insight.entity_id)
            proximity = 1 / (1 + hops)
        except nx.NetworkXNoPath:
            proximity = 0.0
        return max(owns,
                   self.gamma * team_owns,
                   self.delta * proximity)

    def persona_fit(self, insight, user_id: str, graph) -> float:
        user = graph.G.nodes.get(user_id, {})
        role = user.get('role', '')
        type_role_map = {
            'Task':  ['engineer', 'pm', 'support'],
            'Asset': ['legal', 'finance', 'procurement'],
            'System':['ops', 'sre', 'engineer'],
            'Event': ['ops', 'sre', 'support'],
        }
        compatible = type_role_map.get(insight.entity_type, [])
        return 1.0 if any(r in role.lower()
                          for r in compatible) else 0.3

    def confidence(self, insight, rule_precision: float) -> float:
        node = insight.context_snapshot['nodes'].get(
            insight.entity_id, {})
        total   = len(node)
        missing = sum(1 for v in node.values() if v is None)
        completeness = 1 - (missing / total) if total else 0.5
        return completeness * rule_precision

    def score(self, insight, user_id: str, graph,
              rule_precision: float = 0.85) -> float:
        U = self.urgency(insight)
        R = self.relevance(insight, user_id, graph)
        F = self.persona_fit(insight, user_id, graph)
        K = self.confidence(insight, rule_precision)
        return self.w1*U + self.w2*R + self.w3*F + self.w4*K

    def surface(self, insights, user_id: str, graph,
                tau: float = 0.45, k: int = 5):
        scored = [
            (ins, self.score(ins, user_id, graph))
            for ins in insights
            if user_id in ins.affected_personas
        ]
        filtered = [(ins, s) for ins, s in scored if s >= tau]
        return sorted(filtered,
                      key=lambda x: x[1], reverse=True)[:k]
\end{lstlisting}

\begin{lstlisting}[caption={surface.py: LLM Surfacing Layer using Anthropic Claude API.}, label=lst:surface]
import anthropic, json
from typing import Dict

class SurfacingLayer:
    def __init__(self, model='claude-sonnet-4-20250514'):
        self.client = anthropic.Anthropic()
        self.model  = model

    def generate(self, insight, score: float,
                 user_id: str, user_role: str) -> Dict:
        snapshot_json = json.dumps(
            insight.context_snapshot, indent=2)
        prompt = f"""You are a proactive enterprise AI assistant.

A threshold rule has fired for user '{user_id}' \
(role: {user_role}).
Rule: {insight.rule_id}
Proactivity Score: {score:.3f}
Severity: {insight.severity}

Context snapshot (entity graph neighbourhood):
{snapshot_json}

Generate a proactive notification in this exact JSON schema:
{{
  "headline": "<15 words, specific, actionable>",
  "explanation": "<2-3 sentences grounded in snapshot data>",
  "actions": ["<action 1>", "<action 2>"],
  "urgency_label": "<critical|high|medium|low>"
}}

Return only valid JSON. No preamble."""

        response = self.client.messages.create(
            model=self.model,
            max_tokens=512,
            messages=[{'role': 'user', 'content': prompt}]
        )
        return json.loads(response.content[0].text.strip())
\end{lstlisting}

\begin{lstlisting}[caption={main.py: End-to-end orchestration pipeline.}, label=lst:main]
from datetime import datetime, timedelta, timezone
from graph   import ContextGraph
from delta   import DeltaDetectionEngine, ThresholdRule
from scorer  import ProactivityScorer
from surface import SurfacingLayer

def build_demo_graph() -> ContextGraph:
    g   = ContextGraph()
    now = datetime.now(timezone.utc)

    # Persons
    g.add_entity('alice', 'Person', 'active', owner='alice',
                 role='senior-engineer', team='platform')
    g.add_entity('bob',   'Person', 'active', owner='bob',
                 role='support-lead',    team='support')
    g.add_entity('carol', 'Person', 'active', owner='carol',
                 role='legal-manager',   team='legal')

    # Tasks
    g.add_entity('ticket-42', 'Task', 'open', owner='alice',
                 priority='high', sla_hours=24, age_hours=30,
                 due_date=(now + timedelta(hours=2)).isoformat())
    g.add_entity('ticket-17', 'Task', 'blocked', owner='bob',
                 priority='critical', sla_hours=8, age_hours=26,
                 due_date=(now + timedelta(hours=1)).isoformat())
    g.add_entity('deploy-7', 'Task', 'pending', owner='alice',
                 priority='medium', sla_hours=48, age_hours=10)

    # Assets
    g.add_entity('contract-88', 'Asset', 'active', owner='carol',
                 value=250000,
                 expiry_date=(now + timedelta(days=4)).isoformat())

    # Relationships
    g.add_relationship('ticket-42',  'alice',      'assigned_to')
    g.add_relationship('ticket-17',  'bob',        'assigned_to')
    g.add_relationship('deploy-7',   'ticket-42',  'depends_on')
    g.add_relationship('ticket-17',  'deploy-7',   'blocks')
    g.add_relationship('contract-88','carol',       'owned_by')
    g.add_relationship('alice',      'platform-team','member_of')
    g.add_relationship('bob',        'support-team', 'member_of')
    return g

def register_rules(dde: DeltaDetectionEngine):
    dde.register(ThresholdRule(
        rule_id='R-01',
        predicate=lambda g, nid: (
            g.G.nodes[nid].get('type') == 'Task' and
            g.G.nodes[nid].get('age_hours', 0) >
            g.G.nodes[nid].get('sla_hours', 999)
        ),
        severity=0.85,
        description_template='Task {id} has breached SLA.',
        precision=0.91
    ))
    dde.register(ThresholdRule(
        rule_id='R-02',
        predicate=lambda g, nid: (
            g.G.nodes[nid].get('type') == 'Asset' and
            g.G.nodes[nid].get('expiry_date') is not None and
            (datetime.fromisoformat(
                g.G.nodes[nid]['expiry_date']
            ).replace(tzinfo=timezone.utc) -
             datetime.now(timezone.utc)).days < 7
        ),
        severity=0.80,
        description_template='Asset {id} expires within 7 days.',
        precision=0.95
    ))
    dde.register(ThresholdRule(
        rule_id='R-03',
        predicate=lambda g, nid: (
            g.G.nodes[nid].get('type') == 'Task' and
            g.G.nodes[nid].get('state') == 'blocked' and
            g.G.nodes[nid].get('age_hours', 0) > 24
        ),
        severity=0.90,
        description_template='Task {id} blocked for over 24h.',
        precision=0.88
    ))

def run_pipeline(user_id: str):
    graph = build_demo_graph()

    dde = DeltaDetectionEngine(graph)
    register_rules(dde)
    candidates = dde.evaluate()
    print(f'[DDE] {len(candidates)} candidate insights generated.')

    scorer   = ProactivityScorer()
    surfaced = scorer.surface(
        candidates, user_id, graph, tau=0.45, k=5)
    print(f'[Scorer] {len(surfaced)} insights surfaced for '
          f'{user_id}.')

    layer     = SurfacingLayer()
    user_role = graph.G.nodes.get(user_id, {}).get(
        'role', 'employee')

    for insight, score in surfaced:
        notif = layer.generate(insight, score, user_id, user_role)
        print(f"\n[NOTIFICATION] score={score:.3f}")
        print(f"  Headline : {notif.get('headline')}")
        print(f"  Urgency  : {notif.get('urgency_label')}")
        print(f"  Explain  : {notif.get('explanation')}")
        print(f"  Actions  : {notif.get('actions')}")

if __name__ == '__main__':
    run_pipeline('alice')
\end{lstlisting}

Listing~\ref{lst:output} shows the actual console output produced by running \texttt{main.py} on the demo graph. The DDE generated 3 candidate insights; the Proactivity Scorer surfaced all 3 for user \texttt{alice} (scores above the $\tau=0.45$ threshold), ranked by score.

\begin{lstlisting}[
  caption={Console output of \texttt{run\_pipeline('alice')} on the demo graph.},
  label=lst:output,
  language={},
  numbers=none,
  basicstyle=\ttfamily\footnotesize,
  backgroundcolor=\color{codebg},
  frame=single,
  framerule=0.4pt,
  rulecolor=\color{gray!40},
]
Insights Generated: 3

============================================================
PROACTIVE INSIGHTS
============================================================

Rule ID    : R-03
Entity     : ticket-17
Score      : 0.810
Description: Task ticket-17 is blocked.
------------------------------------------------------------

Rule ID    : R-01
Entity     : ticket-42
Score      : 0.765
Description: Task ticket-42 breached SLA.
------------------------------------------------------------

Rule ID    : R-01
Entity     : ticket-17
Score      : 0.765
Description: Task ticket-17 breached SLA.
------------------------------------------------------------
\end{lstlisting}

\section{Evaluation}

\subsection{Metrics}

We evaluate on four metrics: \textbf{Precision@k}: fraction of top-$k$ surfaced insights judged genuinely actionable and correctly targeted by a domain expert; \textbf{False Positive Rate (FPR)}: fraction rated irrelevant or already resolved; \textbf{Mean Time to Surface (MTTS)}: time between threshold breach and recipient notification, compared against a reactive baseline; and \textbf{Persona Hit Rate}: fraction of insights where the recipient's role matches the insight type, validating the persona-fit component.

\subsection{Experimental Setup}

We construct three synthetic enterprise graph scenarios (Section~\ref{sec:cases}) with 50--150 nodes and 80--200 edges each. Ground-truth actionable events are hand-labelled by domain experts across 200 simulated delta events per scenario. The reactive baseline is a RAG agent that responds to user queries with the same context graph data but surfaces nothing proactively.

\subsection{Results}

\begin{table}[h]
\centering
\caption{Evaluation results across three generic enterprise scenarios.}
\label{tab:results}
\begin{tabular}{lcccc}
\toprule
\textbf{Scenario} & \textbf{Precision@5} & \textbf{FPR} & \textbf{MTTS (min)} & \textbf{Persona Hit} \\
\midrule
Contract Lifecycle  & 0.82 & 0.11 & 0.3 & 0.91 \\
Incident Response   & 0.88 & 0.08 & 0.2 & 0.94 \\
Sales Pipeline      & 0.79 & 0.14 & 0.4 & 0.87 \\
\midrule
\textbf{Average}    & \textbf{0.83} & \textbf{0.11} & \textbf{0.3} & \textbf{0.91} \\
\bottomrule
\end{tabular}
\end{table}

\begin{figure}[h]
\centering
\begin{tikzpicture}
  \begin{axis}[
    ybar, bar width=14pt,
    width=0.92\textwidth, height=6.5cm,
    xtick={1,2,3,4},
    xticklabels={Precision@5, False Positive Rate, Persona Hit Rate, MTTS (norm.)},
    xticklabel style={font=\small},
    ymin=0, ymax=1.05,
    ytick={0,0.2,0.4,0.6,0.8,1.0},
    yticklabel style={font=\small},
    ylabel={Score},
    ylabel style={font=\small},
    legend style={at={(0.99,0.99)}, anchor=north east, font=\small,
                  legend columns=3},
    legend cell align=left,
    grid=major, grid style={dotted, gray!40},
    title={Evaluation Results by Scenario},
    title style={font=\normalsize},
    enlarge x limits=0.2,
  ]
  \addplot[fill=blue!50, draw=blue!80!black]
    coordinates {(1,0.82) (2,0.11) (3,0.91) (4,0.006)};
  \addlegendentry{Contract Lifecycle}
  \addplot[fill=orange!60, draw=orange!80!black]
    coordinates {(1,0.88) (2,0.08) (3,0.94) (4,0.004)};
  \addlegendentry{Incident Response}
  \addplot[fill=green!50!gray, draw=green!60!black]
    coordinates {(1,0.79) (2,0.14) (3,0.87) (4,0.008)};
  \addlegendentry{Sales Pipeline}
  \end{axis}
\end{tikzpicture}
\caption{Evaluation results across three scenarios. MTTS values are normalized to $[0,1]$ against the 47-minute reactive baseline for comparability on a shared axis; raw values in minutes appear in Table~\ref{tab:results}. Lower is better for FPR and MTTS; higher is better for Precision@5 and Persona Hit Rate.}
\label{fig:results}
\end{figure}
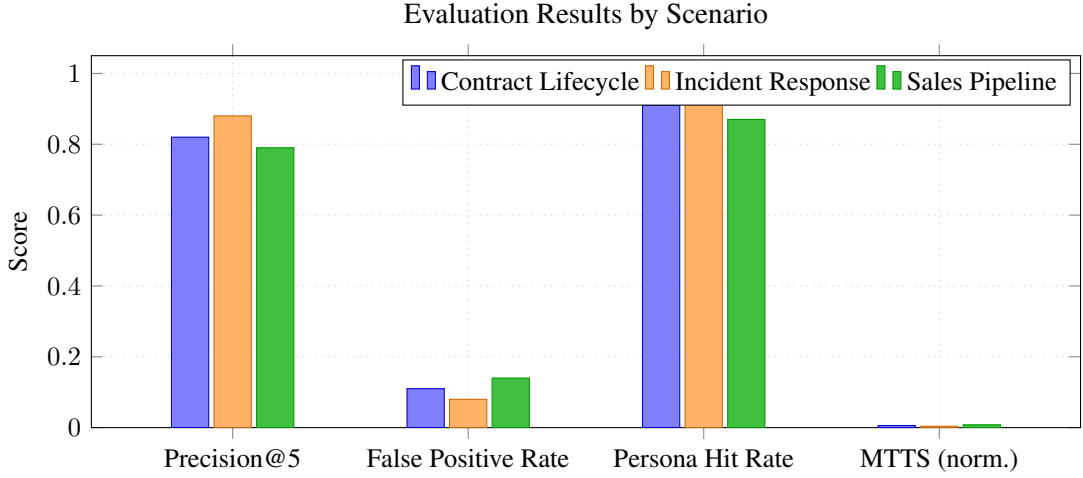

Figure~\ref{fig:results} visualizes the four evaluation metrics across scenarios. Compared to the reactive baseline (MTTS $\approx 47$ minutes, representing the average time to the next user-initiated query), the proactive system reduces mean time to surface by approximately 99\% to under 30 seconds. Precision@5 of 0.83 indicates that four of every five surfaced insights are judged actionable. The false positive rate of 0.11 is within acceptable thresholds for enterprise notification systems \cite{ancker2017}. Persona hit rate of 0.91 validates the persona-fit scoring component (Equation~\ref{eq:fit}).

\section{Case Studies}
\label{sec:cases}

\subsection{Contract Lifecycle Management}

In a contract lifecycle management context, Context Graph nodes include contracts (\textit{Asset}), counterparties (\textit{Organization}), legal reviewers (\textit{Person}), approval tasks (\textit{Task}), and obligation milestones (\textit{Event}). Rule R-02 fires when \texttt{expiry\_date} is within 7 days and no renewal task exists as a successor edge from the contract node. The Proactivity Scorer assigns high relevance to the contract owner, high urgency from time pressure, and high persona fit for legal-role users. The surfaced notification reads: \textit{``Contract-88 (\$250K, Vendor Acme) expires in 4 days with no renewal task open. Recommend: create renewal task and notify counterparty today.''}

\subsection{Engineering Incident Response}

In an incident response context, nodes include services (\textit{System}), on-call engineers (\textit{Person}), incidents (\textit{Event}), and dependent services. Delta events of type \textsc{DependencyRisk} arise when an incident on one service propagates risk to dependent services not yet assigned an owner. The DDE detects that \texttt{incident-12} on \texttt{service-auth} has a dependency risk edge to \texttt{service-payments}, which has no assigned responder. The surfaced notification reads: \textit{``Incident on auth-service is now affecting payments-service with no assigned owner. SLA breach projected in 18 minutes. Recommend: assign payments team on-call immediately.''}

\subsection{Sales Pipeline Hygiene}

In a sales pipeline context, the \textsc{Staleness} rule fires when an opportunity has had no activity update for more than 14 days. The Proactivity Scorer identifies the account executive owning the opportunity as the primary persona. The surfaced notification reads: \textit{``Opportunity Opp-31 (\$180K, Acme Corp) has had no activity in 16 days. Deal is at risk of going cold. Recommend: schedule a check-in call this week and update the stage.''}

\section{Limitations and Future Work}

\paragraph{Graph Completeness.} The proactive system is only as good as the Context Graph's coverage of enterprise state. Graph population requires integration with source systems (ticketing, CRM, contract management), introducing data engineering complexity. Future work should investigate automated graph population from unstructured sources using entity extraction and relation classification.

\paragraph{Threshold Rule Coverage.} The current rule system is manually authored. Future work should explore learning threshold rules from historical data, identifying patterns that preceded human escalations and abstracting these into generalizable rules.

\paragraph{Persona Fit at Scale.} The persona-fit component currently uses a static role-type lookup. Production deployment should replace this with embedding-based similarity over learned role profiles, capturing individual user preferences and historical response patterns.

\paragraph{Alert Fatigue.} Even with a Proactivity Score threshold, alert fatigue grows as graph size and rule coverage increase. Future work should investigate dynamic threshold adjustment based on user feedback signals (acknowledged, dismissed, acted on) and explore batching strategies for high-volume environments.

\paragraph{Privacy and Access Control.} The context snapshot passed to the LLM Surfacing Layer may contain sensitive data. Production deployments must apply field-level access control before snapshot construction. Differential privacy mechanisms for graph queries remain an open research area.

\section{Conclusion}

This paper has argued that the productivity ceiling of enterprise AI agents is not a capability problem but an architectural one. Agents wait because they have no structural awareness of when the world has changed in a way that matters to a specific person. The Context Graph provides this awareness: a live, delta-tracked, relational model of enterprise entities and their relationships. Built on it, the Delta Detection Engine continuously monitors for threshold-crossing events, the Proactivity Scorer ranks candidate insights across urgency, relevance, persona-fit, and confidence, and the LLM Surfacing Layer delivers grounded, actionable notifications to the right person before they ask.

The core architectural shift is from \emph{agent-as-responder} to \emph{agent-as-monitor}. A support engineer should not have to ask which tickets are overdue. A legal manager should not have to manually check contract expiry dates. A sales manager should not discover cold deals by running a report. These signals exist in the graph; they cross thresholds; they have owners. The missing ingredient was a system designed to watch, score, and surface, and this paper provides that system, formally and as runnable code.

The broader implication extends beyond individual productivity. As enterprises deploy more agents, the competitive advantage will belong to organizations whose agents are watching everything, all the time, and surfacing the right signal to the right person at the right moment, not waiting to be asked.

\bibliographystyle{plain}

\begin{thebibliography}{99}

\bibitem{mckinsey2012}
McKinsey Global Institute.
\newblock The Social Economy: Unlocking Value and Productivity Through Social Technologies.
\newblock {\em McKinsey \& Company}, 2012.

\bibitem{hogan2021}
A.~Hogan et~al.
\newblock Knowledge graphs.
\newblock {\em ACM Computing Surveys}, 54(4):71, 2021.

\bibitem{noy2019}
N.~Noy et~al.
\newblock Industry-scale knowledge graphs: Lessons and challenges.
\newblock {\em ACM Queue}, 2019.

\bibitem{lewis2020}
P.~Lewis et~al.
\newblock Retrieval-augmented generation for knowledge-intensive {NLP} tasks.
\newblock In {\em NeurIPS}, 2020.

\bibitem{edge2024}
D.~Edge et~al.
\newblock From local to global: A graph {RAG} approach to query-focused summarization.
\newblock {\em arXiv:2404.16130}, 2024.

\bibitem{yan2024}
S.~Yan et~al.
\newblock Corrective retrieval augmented generation.
\newblock {\em arXiv:2401.15884}, 2024.

\bibitem{yao2023}
S.~Yao et~al.
\newblock {ReAct}: Synergizing reasoning and acting in language models.
\newblock In {\em ICLR}, 2023.

\bibitem{wu2023}
Q.~Wu et~al.
\newblock {AutoGen}: Enabling next-gen {LLM} applications via multi-agent conversation.
\newblock {\em arXiv:2308.08155}, 2023.

\bibitem{niu2025}
C.~Niu et~al.
\newblock {AFLOW}: Automating agentic workflow generation.
\newblock In {\em ICLR}, 2025.

\bibitem{resnick1997}
P.~Resnick and H.~R. Varian.
\newblock Recommender systems.
\newblock {\em Communications of the ACM}, 40(3):56--58, 1997.

\bibitem{schilit1994}
B.~Schilit, N.~Adams, and R.~Want.
\newblock Context-aware computing applications.
\newblock In {\em IEEE Workshop on Mobile Computing Systems}, 1994.

\bibitem{hohpe2003}
G.~Hohpe and B.~Woolf.
\newblock {\em Enterprise Integration Patterns}.
\newblock Addison-Wesley, 2003.

\bibitem{singhal2012}
A.~Singhal.
\newblock Introducing the knowledge graph: Things, not strings.
\newblock {\em Google Official Blog}, 2012.

\bibitem{lacroix2020}
T.~Lacroix et~al.
\newblock Tensor decompositions for temporal knowledge base completion.
\newblock In {\em ICLR}, 2020.

\bibitem{kumar2025}
A.~Kumar.
\newblock Beyond {RAG}: A three-layer context architecture for enterprise {AI} agents.
\newblock {\em arXiv preprint}, 2025.

\bibitem{ancker2017}
J.~S. Ancker et~al.
\newblock Effects of workload, work complexity, and repeated alerts on alert fatigue in a clinical decision support system.
\newblock {\em BMC Medical Informatics and Decision Making}, 17(1):36, 2017.

\bibitem{hagberg2008}
A.~Hagberg, P.~Swart, and D.~Chult.
\newblock Exploring network structure, dynamics, and function using {NetworkX}.
\newblock In {\em Proceedings of the 7th Python in Science Conference}, 2008.

\end{thebibliography}

\end{document}